\documentclass[10pt,letterpaper]{article}
\usepackage{cogsci}
\usepackage{lmodern}
\cogscifinalcopy 
\usepackage{pslatex}
\usepackage{float}
\usepackage{stmaryrd}
\usepackage{amsmath}
\usepackage{amsfonts}
\usepackage{simplebnf}
\usepackage{subcaption}
\usepackage{CJKutf8}
\usepackage{booktabs}
\usepackage{semantic}
\usepackage[
    backend=biber,
    style=apa,
    natbib=true,
    doi=false,
    isbn=false,
    url=false,
]{biblatex}
\title{A Computational Cognitive Model for Processing \\Repetitions of Hierarchical Relations}
 
\author{
    {\large \bf Zeng Ren (zeng.ren@epfl.ch)} \\
  \AND 
    {\large \bf Xinyi Guan (xinyi.guan@epfl.ch)} \\
  \AND
    {\large \bf Martin Rohrmeier (martin.rohrmeier@epfl.ch)} \medskip\\ 
    Digital and Cognitive Musicology Lab,\\
    École Polytechnique Fédérale de Lausanne\\
    1015 Lausanne, Switzerland
  }

\begin{document}

\maketitle
    
\begin{abstract}
Patterns are fundamental to human cognition, enabling the recognition of structure and regularity across diverse domains. 
In this work, we focus on the \emph{structural repeats}, patterns that arise from the repetition of hierarchical relations within sequential data, and develop a candidate computational model of how humans detect and understand such structural repeats. 
Based on a weighted deduction system, our model infers the minimal generative process of a given sequence in the form of a Template program, a formalism that enriches the context-free grammar with repetition combinators. Such representation efficiently encodes the repetition of sub-computations in a recursive manner.
As a proof of concept, we demonstrate the expressiveness of our model on short sequences from music and action planning. The proposed model offers broader insights into the mental representations and cognitive mechanisms underlying human pattern recognition.

\end{abstract}

\section{Introduction}

    Patterns are ubiquitous across domains, arising from the repetition of invariance—whether explicit or abstract—among their instances. 
    From recognizing recurring architectural structures in cityscapes to identifying parallel syntactic forms in poetry, humans effortlessly detect and utilize patterns despite variations in their individual elements. This capacity extends to diverse domains: we discern shared narrative structures across novels, films, and music, and we uncover universal laws governing phenomena as distinct as ocean currents and atmospheric dynamics.  
    Moreover, humans can abstract patterns of patterns, as seen in the application of category theory to mathematics and programming.
    These examples illustrate our remarkable ability to not only detect abstract repetitions but also productively employ them in navigating both the physical and mental worlds.
    Such capacity is believed to be a fundamental aspect of human cognition \parencite{watanabe1985pattern,margolis1987patterns,pavlidis2013structural,pomiechowska2024compositionality}.

\begin{figure*}[t]
        \begin{subfigure}[b]{0.5\linewidth}
            \centering
            \includegraphics[width=0.9\linewidth]{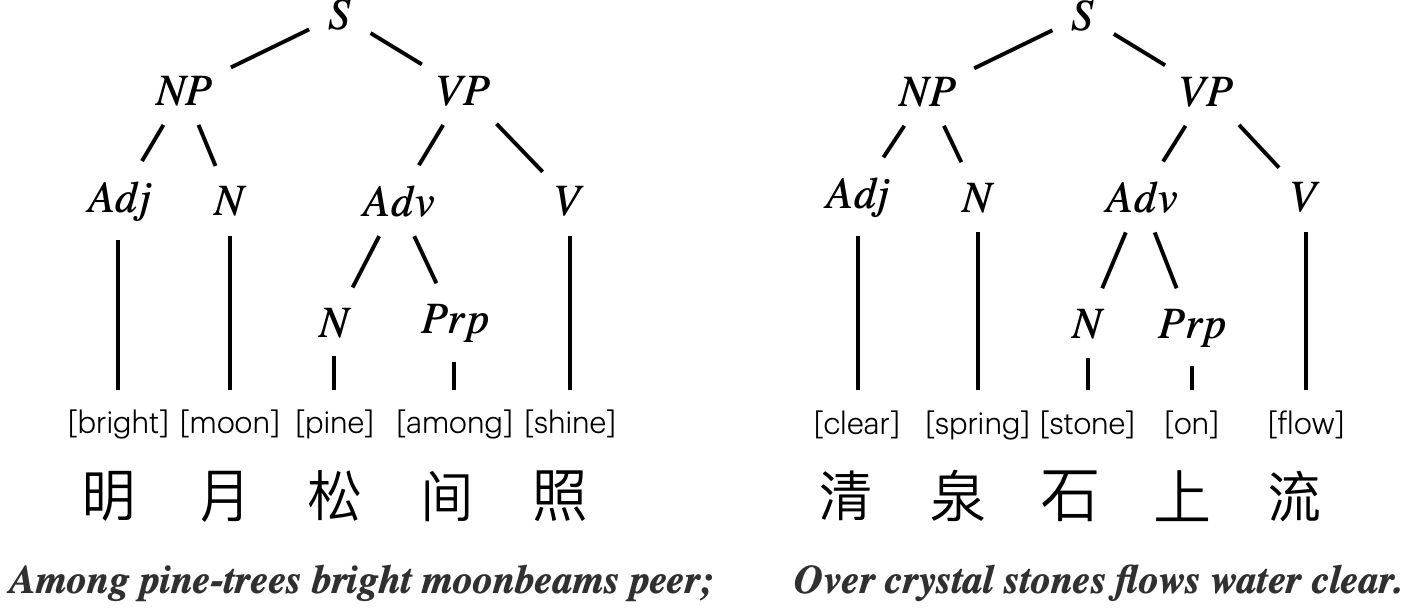} 
            
            \caption{
            }
            \label{fig:poetry}
            
        \end{subfigure}
        \begin{subfigure}[b]{0.5\linewidth}
            \centering
            \includegraphics[width=0.9\linewidth]{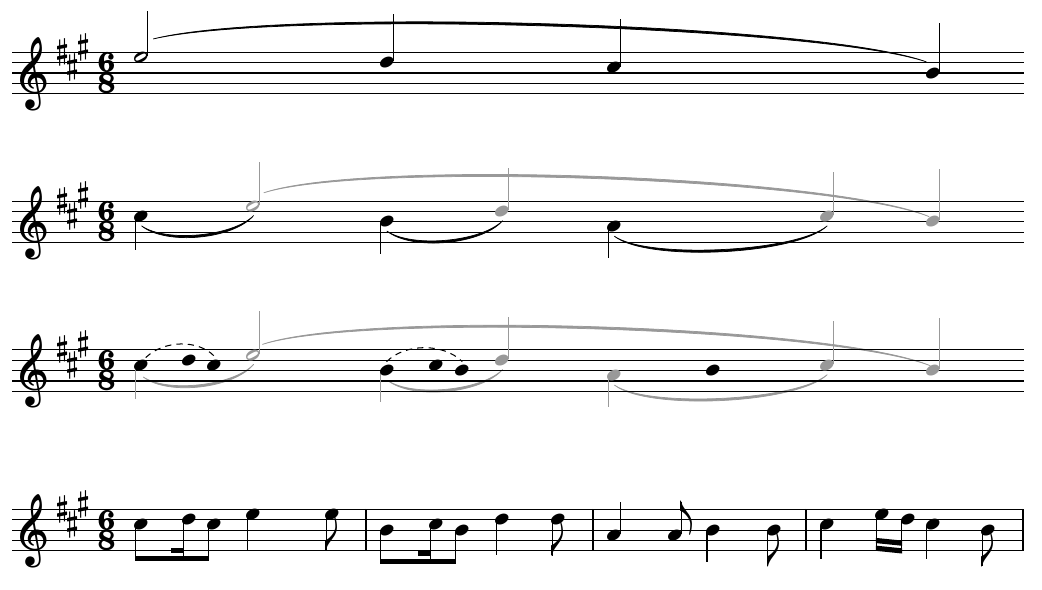}
            \caption{
            }
            \label{fig:music}
        \end{subfigure}
        
        \begin{subfigure}{\linewidth}
            \centering
            \includegraphics[width=1\linewidth]{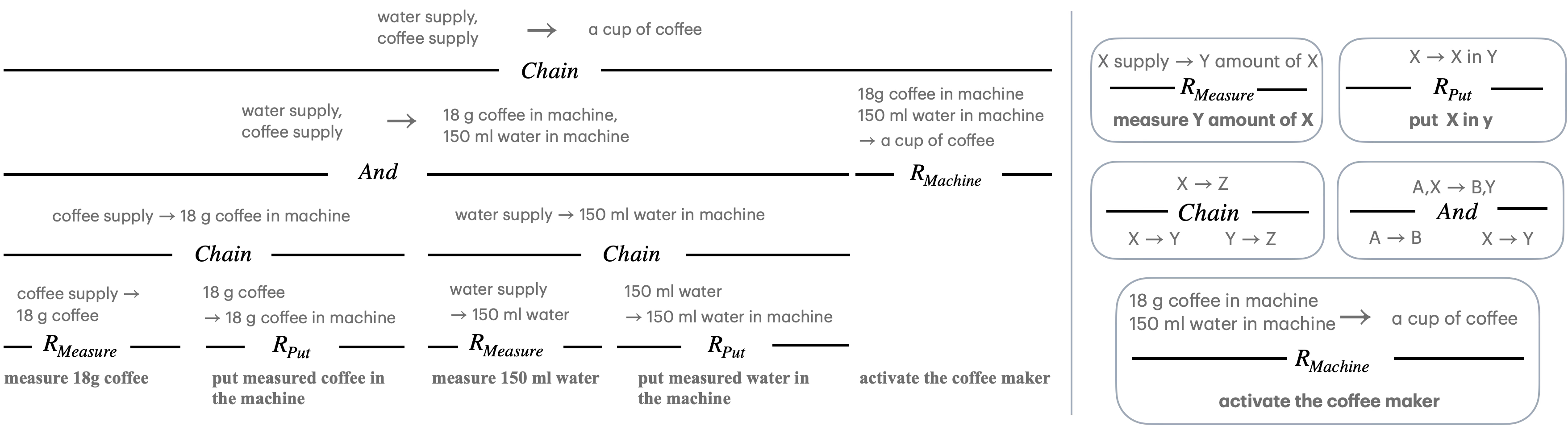} 
            \caption{
            }
            \label{fig:coffee}
        \end{subfigure}
        \caption{Three examples of structural repetition across domains demonstrates the kinds of computation involved in structural repeat. (\ref{fig:poetry}) An excerpt from the Tang poetry ``\begin{CJK*}{UTF8}{gbsn}\text{山居秋暝}\end{CJK*}'' written by Wang Wei (699–761), translated by Xu Yuanchong, exhibits parallel syntactic structure (repetition of complete computation). (\ref{fig:music}) A melodic reduction of the opening theme in K331 (mm. 1-4) shows mm. 1-3 share the same underlying generative process up to a certain point (repetition of suspended computation). (\ref{fig:coffee}) The hierarchical action planning involved in making coffee reveals polymorphic relations as the basis of structural repeats.}
        \label{fig: examples of structural repeat}
    \end{figure*}

    Here, we focus on patterns that arise from the repetition of hierarchical relations within sequential data, which we term \emph{structural repeats}. Unlike surface-level repetitions, structural repeats are not directly observable in the sequence itself but are inferable from the underlying generative process. Fig. \ref{fig: examples of structural repeat} illustrates examples of structural repeats in diverse domains, including poetry, action planning, and music. Despite surface-level differences in each of the sequences, humans can perceive and appreciate the repetition of their underlying ``logic.''
    
    How do humans infer the repetition of abstract patterns? Computational modeling in cognitive science serves as an important means to address such questions by providing explicit and testable mechanisms for how the mind processes information \parencite{lake2017building, gershman2015computational, sun2009theoretical, sun2005levels}. In particular, generative models have been widely used to study sequences usually in form of (probabilistic) context-free grammar \parencite{jelinek1992basic}. Recent advances in program induction have further highlighted the connection between learning and the inference of generative programs that produce observed outputs \parencite{chater2013programs, rothe2017question, lake2015human,ellis2021dreamcoder,schmid2011inductive,rule2024symbolic}, as emphasized in the 2018 CogSci workshop on ``Learning as Program Induction'' \parencite{Bramley2018Induction}. Building on these approaches, we study structural repeats using program induction methodologies guided by the principle of minimal description length \parencite{grunwald2007minimum}.
  
    Our work adopts the following strategy: we first analyze the properties of mental processes involved in pattern recognition and use these insights to derive a mathematical model—a mini-programming language—that defines a hypothesis space of programs capable of generating observed sequences.
    Following the idea of ``learning as program induction,'' we develop a weighted deduction algorithm \parencite{pereira1987prolog, sikkel1997parsing, shieber1995principles, nederhof2003weighted, Jason2023time} to infer programs that evaluate to the observed sequence.
    While many candidate programs can produce the same output, humans do not consider all possibilities equally plausible. To address this, we use Minimum Description Length (MDL) \parencite{grunwald2007minimum} as a guiding principle to rank and disambiguate among programs, favoring those that offer the most compact representations.
    Finally, we assess the plausibility of the minimal program. 
    Our computational model, together with its implementation, provides a candidate theory and explicit descriptions on how the human minds recognize and process abstract relational repeats in sequential data.
  
\subsection{Characterizing structural repetition}
    The examples of poetry (Fig. \ref{fig:poetry}), music (Fig. \ref{fig:music}), and hierarchical planning (Fig. \ref{fig:coffee}) progressively illustrate the complexities of structural repetition and the computational mechanisms required to process them.
    In poetry, parallel syntactic structures demand a hierarchical interpretation of sequences and the ability to identify the repeated substructures, such as derivation trees. 
    In the musical example, repeated computations can involve incomplete structures or ``holes,'' where repetitions are not exact subtrees but partial ones. 
    A top-down derivation reveals that notes $\hat{5}-\hat{4}-\hat{3}$ are first elaborated in the same way using ascending thirds (second row in Fig.\ref{fig:music}). This parallel construction then changes when the first two bars are further elaborated using neighboring motion whereas the third bar uses a passing motion (third row). 
    Fig. \ref{fig:coffee} presents a hierarchical planning of actions involved in the task of making a cup of coffee from its ingredients. Notice that the actions involving preparing coffee ground and preparing water are ``repeated'' not in the literal sense but in terms of the relational structure of the underlying tasks. This example highlights the role of relations as the basic repeating units, emphasizing the repetition of computational processes (e.g., polymorphic production rules) rather than the input/output states of the computation (e.g., (non)terminals symbols).

    To summarize, we characterize structural repetition of hierarchical relations by two key properties. First, a single relation can manifest in multiple forms, similar to logical clauses involving meta-variables. Second, these relations can be  constructed inductively via composition and duplications. Such construction process further entails the ability to express a) incomplete/suspended computations and b) various ways to repeat (the bindings of function variables in a composition expression).

\subsection{Related formalisms}
    The pattern language \parencite{angluin1979finding} provides a formal model for finding repetitive patterns in sequences such as $aXbcX$ where $a, b, c$ are terminal symbols and the variable $X$ can be substituted for other patterns. This approach allows for capturing varied repeats of symbols and generative processes with ``holes'' that must contain shared structures. However, it only concerns the symbols rather than the \emph{relations} among the symbols. Consequently, this makes it not applicable for the cases such as our musical example (see Fig. \ref{fig:music}) where repeating entities are not the notes themselves but their relations.   

    Formalisms for generative processes, particularly those involving shared computations, have also been developed. 
    A prominent example is Tree-Adjoining Grammar (TAG), a mildly context-sensitive formalism widely used in linguistics \parencite{joshi1997tree,joshi1987introduction,ferreira2013syntax}. 
    TAG constructs new structures by reusing a set of predefined computations(``initial trees'' and ``auxiliary trees'') that potentially contain holes so they may be treated as functions whose input are trees. In addition, fragment grammar \parencite{odonnell2009fragment} infers the distribution of such composed computations directly from the observed sequential data, and provides a plausible explanation of how human dynamically manage such set of stored computations.

    One potential approach to integrate tree-based languages with pattern languages \parencite{angluin1979finding} is to use trees with variables, such as let-bindings in expressions.
    However, maintaining variable bindings within deeply nested computations is cognitively unfavorable \parencite{sweller1988cognitive, halford1998processing, cowan2001magical}.
    An alternative is combinatorial calculus, a variable-free framework proposed as a model for the language of thought \parencite{dechter2018using}. As a result of its generality, its corresponding program space (Turing-complete computations) is unnecessarily large for our specific use case (modeling structural repeat); inferring such a program (and finding the smallest one) poses a difficult and potentially intractable task. Thus, we look for a formalism that captures structural repetitions using a variable-free representation and integrating features of tree language with pattern language in a minimal way.

    Ren et al. \parencite*{ren2024formal} introduced the \emph{Template} model to capture structural repeats in annotated parse trees of musical data (jazz harmony), proposing a set of primitive combinators called ``meta-rules'' to express repetitions within a tree structure.
    For example, the combinator $\langle\_ \ 0 \rangle$ corresponds to the higher-order function $\lambda g.\lambda f. \lambda x. \lambda y.\  g\  (f\  x )\  (f\  y)$, \footnote{Following the convention in lambda calculus, we express function applications  as $f\ x \ y$ rather than $f(x,y)$.} which expresses repeated sibling computation (each contains one hole that can be freely instantiated). It also includes combinators with parent/child repeat (single step of recursion) such as $\langle \star\  \_ \rangle = \lambda g.\lambda f.\lambda x.\lambda y.\lambda z.\ g\ (g \ x\ y)\ (f\ z)$.
    Our work extends the Template model in two key ways: while the former model processes structured parse trees, our approach operates directly on raw sequential inputs, and instead of using tree compression heuristics to find the smallest template, we provide an exact algorithm for optimal template discovery.

\section{Template program: The computational model}

    \begin{figure}
        \centering
        \includegraphics[width=\linewidth]{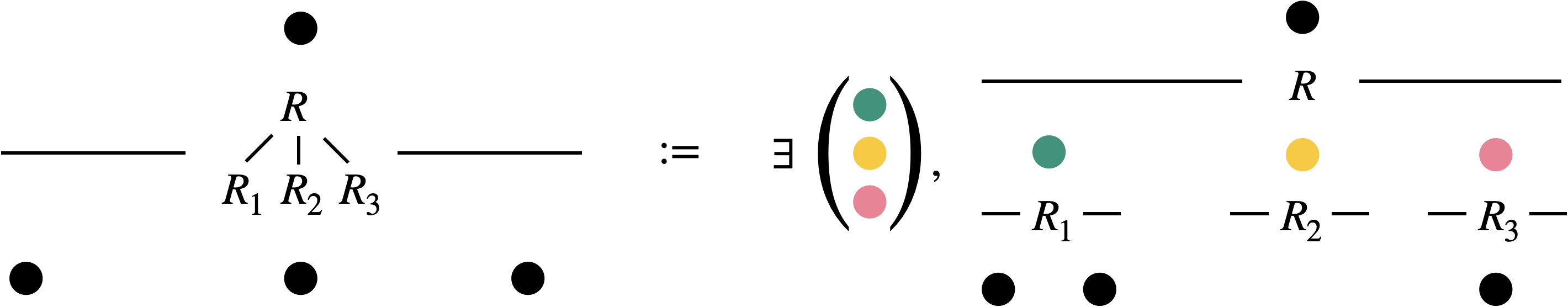}
        \caption{Visual illustration on how hierarchical relations compose. $R$ has arity 3, $R_1$ has arity 2, $R_2$ has arity 0, $R_3$ has arity 1, and the composed relation $R \circ (R_1 \otimes R_2 \otimes R_3)$ has arity $2+0+1=3$.}
        \label{fig:relation_compose}
    \end{figure}

    We define a hierarchical relation $R$ as a $n+1$-ary relation of the form $xR x_1\dots x_n$ where the $x$ is called the head symbol and the $x_i$s are called child symbols. Relations can be composed by unifying the head of a relation with a child of another as shown in Fig. \ref{fig:relation_compose}.
    Relations generalize functions by allowing bi-directional and non-deterministic mapping. A production rule $P 
    := X \to Y\  Z$ of a context-free grammar is a special case of a hierarchical relation that only has one possible instantiation.     
    A production function may allow multiple instantiations, and the child symbols (right-hand-side) $Y$ and $Z$ are uniquely determined by the head symbol (left-hand-side) $X$. In contrast, an inference rule of a deductive system $\frac{B \quad C}{A}$ is a hierarchical relation where the head symbol (conclusion) $A$ is uniquely determined by the child symbols (premises) $B$ and $C$.

\subsection{Language Specification}

\paragraph{Expression Syntax} Given a set of relation $R$ from $X^n$ to $X$, we define the syntax of the Template program as:\\

\begin{bnf}
$e$ : \textsf{Expr} ::=
| \texttt{Id} : identity relation
| \texttt{Pure $r$} : relation
| \texttt{Rep $e \ m \ [e]$} : repetition
;;
$m$ : \textsf{Combinator} ::= 
| \texttt{$[s]$} : 
;;
$s$ : \textsf{} ::= 
| \texttt{$\_$} : free variable
| \texttt{$i\in\mathbb{N}$} : sibling repeat 
| \texttt{$\star$} : recursion
\end{bnf}

\paragraph{Semantics} Template expressions can be interpreted as hierarchical relations. We use $\circ$ to denote sequential composition and $\otimes$ for parallel composition of relations.\footnote{We inherit these notations from category theory, as relations form a Cartesian monoidal category.} 

\begin{align*}
    \llbracket \texttt{Pure $x$} \rrbracket &= x\\ 
    \llbracket \texttt{Rep $t$ $m$ $ts$} \rrbracket &= \llbracket t \rrbracket \circ \bigotimes_{t' \in useRep(t, m, ts)}{\llbracket t' \rrbracket}\\
\end{align*}
The $useRep$ function simply applies the repetition combinator defined in \parencite{ren2024formal} to the parent and child relations. For example, $useRep(t,\langle \_ \ 0\  \star\rangle, [t_1]) = [t_1,t_1,t]$.

\subsection{Hypothesis space for an observed sequence}
    There exists a canonical embedding from the derivation tree of a sequence under a context-free grammar to a template program, where the primitive relations correspond to the production rules of the grammar. 
    In this embedding, every derivation tree is a template program where expressions of the form  \texttt{Rep e m es} are specialized to  $\texttt{Rep (Pure r) $\langle\_\dots   \_\rangle$ es}$, with $r$ representing a production rule and the repetition combinator being trivial repeat (implying no repetition). 
    However, the space of possible template programs is vast. For a tree whose nodes are production rules, each partition of the tree into a root tree and a collection of subtrees corresponds to a distinct template program.  
    Furthermore, the choice of compatible repetition combinator $m$ introduces additional ambiguities. 
    Despite this complexity, there exists a compact template program in this space, and it is this program that we seek to approximate as a plausible mental representation of the abstract repetition patterns in the sequence.

\subsection{Disambiguation based on minimal description}
    A simple and natural way to disambiguate among the numerous possible template programs is to select the smallest one(s). 
    The repetition combinators in a template programs provides a way to factor out repeated computations, thus the template size can be smaller than the sum of its components computation. We define the size of a template program as:
    \begin{align}
        &|\texttt{Id}| &&= 1 \\
        &|\texttt{Pure r}| &&= 1 \\
        &|\texttt{Rep e m es}| &&= |\texttt{e}| + 1 + \sum_{\texttt{e}_i \in \texttt{es}}  |e_i| \label{Eq: repsize}
    \end{align}

    Here, \texttt{Id} represents a trivial template program that simply returns its input, such a template program has size 1. 
    For \texttt{Pure $r$}, which represents a single primitive relation, the size is one, as these primitives can be encoded using a finite list of symbols.
    In the inductive case, the size of the program is naturally defined as the size of its free child programs plus one for the repetition combinator.

\section{Deductive Parsing for Template Program}

    Given a sequence observation, we want to find all possible (or the minimal) template program that explains the sequence. We present our parsing algorithm in the framework of weighted deduction \parencite{pereira1987prolog, sikkel1997parsing, Jason2023time}.\footnote{We can not use a semiring parsing framework \parencite{goodman1999semiring} because it cannot express the weight combination function as dependent on the proofs.} In our specific case, the weight $\mathbb{W}$ is interpreted as the minimum size of a template program. Our  weight aggregator $\oplus$ is the $min$ operator. The weight combination function $\texttt{MergeW}_m$ 
    $$\texttt{MergeW}_m(w,w_1\dots w_n)= w + \sum_{v \in \{w_i | m_i=\texttt{free}\}_{i=1}^{n}} w$$ represents how the minimal size of a template expression can be deduced from those of its evaluated sub-programs in a way that is consistent with Eq.\ref{Eq: repsize}. 
 
    In order to encode a group of computation flows with repetition combinators, we first need to check if the underlying computation flows are equal.
    For this reason, the items in our deduction system must explicitly represent the computation flow. This representation is encoded as a tree $t \in \texttt{Tree}_R$ where the nodes are relations in $R$, and the leaves are either unary relations (such as termination rules in context-free grammars) or a hole symbol (which simulates lambda abstraction).

    To determine if a group of computations can explain a segment of the observed sequence, we will encode which subset of the sequence they can cover. 
    In CYK parsing for context-free grammar, it is sufficient to encode a interval $(i,j)$. The composibility of the two intervals depends on whether or not they share the same boundary. Yet, in our case, since each computation flow may contain holes, which correspond to gaps in the sequence, the items must also encode the spans of these gaps. For each item, there is a region attribute, interval type $\texttt{Type}_I$, that encodes the holes' spans $\{I_i\}_{i=1}^n$ and the overall span $I$ of the computation flow. Such encoding corresponds to a deductive reasoning: ``if span $I_1...I_n$ are explained, the current computation flow can explain $I$.'' From this perspective, the CYK algorithm becomes a special case when $n=0$.
    To decide if a group of computation flows are composable (as in Fig.\ref{fig:relation_compose}), we also need to encode their computation types $\texttt{Type}_X$. A relation $xRx_1x_2..x_n$ will have computation type $x_1 \to x_2 \to \dots \to x_n \to x$.
    
\begin{figure*}[t]
        \centering
        \includegraphics[width=0.8\textwidth]{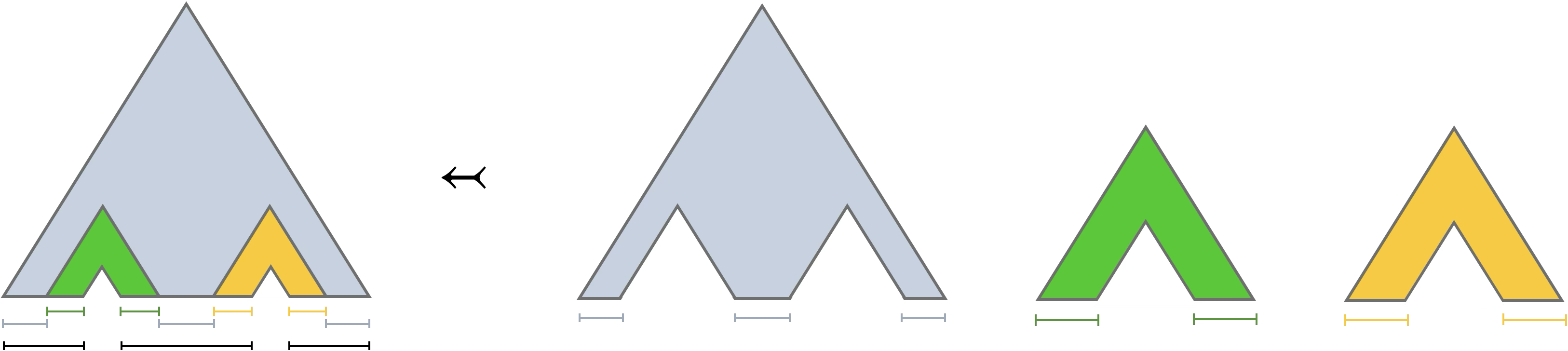}
        \caption{A visual illustration of the inference rule \textsc{Complete-rep}, showing \texttt{mergeT} (composition of the colored shapes) and \texttt{mergeR} (composition of the surface segments). Relation trees are represented by triangles potentially containing holes. This example is a special case where $t$ has two holes (arity = 2) while $t_1$ and $t_2$ each has one hole (arity = 1).   }
        \label{inferenceRuleVis}
    \end{figure*}

    To summarize, each item of our deduction system contains three pieces of information:
    1) the type of its encoded computation 2) its computation flow $t\in \texttt{Tree}_R$ 3) the type of its explainable region (interval type) $\sigma \in \texttt{Type}_I$. There are two ways an item can be initialized in the chart. First is when we infer the non-terminals from terminals using terminating relations (\textsc{Scan-Rel}) and second, when we embed primitive relations (e.g. production rules) as an item (\textsc{Prim-Rel}). The latter approach allows us to recursively build up ``composite production rules'' (computations containing holes). During the deduction process, the inference rule \textsc{complete-rep} (visualized in Fig \ref{inferenceRuleVis}) test whether the computation flow represented by $n+1$ items can be composed as $f\circ \bigotimes[f_1\dots f_n]$ while all satisfying the composibility criteria for its computation type and interval type.
      
    \newcommand{\itemForm}{
        \begin{minipage}{\textwidth}
        $[\tau,t,\sigma] \in \texttt{Type}_X \times\texttt{Tree}_R       \times \texttt{Type}_I $
        \end{minipage}
    }

    \newcommand{\axiom}{
        \begin{minipage}{\textwidth}
        \begin{tabular}{ll}
        \textsc{Scan-Rel:} & $\inference{}{[x,T,(i,i+1)]} xTw_i$ where $T$ is a termination relation \\
        \textsc{Prim-Rel:} & $\inference{}{[\vec x \to x,\ R,\vec I \to I]}xR \vec x \wedge \left(\forall i, |I_i| >0 \right)\wedge \left( I_1 + \dots +  I_n = I\right)$
        \end{tabular}
        \end{minipage}
    }
    
    \newcommand{\inferenceRule}{
        \begin{minipage}{\textwidth}
        \begin{tabular}{ll}
            \textsc{Complete-Rep:} & $\inference{\texttt{$[\tau,t,\sigma]$  $[\tau_1,t_1,\sigma_1] \dots [\tau_n,t_n,\sigma_n]$}} 
            {[\texttt{mergeType}(\tau,\vec{\tau}),\texttt{mergeT}(t,\vec t),\texttt{mergeR}(r,\vec r) ]}
            \texttt{hasRep}_m(t,\vec t) \wedge x'R_t\vec\tau
            $ 
        \end{tabular}

        \end{minipage}
    }
    
    \begin{table*}[t]
        \centering
        \bgroup
        \def\arraystretch{3}%
        \begin{tabular*}{\linewidth}{ll}
        
            \toprule
            \textbf{Sets} & 
            $X := \text{Nonterminals}$ \quad 
            $R := \text{Relations between $X^n$ and $X$}$ \quad
            $I := \text{Intervals in } \mathbb{N}$ \quad 
            $\texttt{Type}_A := A \ |\  A \to \texttt{Type}_A$
            \\
            \textbf{Variables}& 
            $\tau \in \texttt{Type}_X$ \quad  $x\in X$ \quad 
            $t \in \texttt{Tree}_R$ \quad 
            
            $\sigma \in \texttt{Type}_I$ \quad 
            $\vec a = \{a_l\}_{l=1}^n \in A^n$ \quad 
            $i,j \in \mathbb N$ \quad 
            $r \in \texttt{Type}_I $  \quad 
            $v \in \mathbb N$ \\ 
            
            \textbf{Items}& \itemForm \\ 
            \textbf{Axioms}&  \axiom \\ 
            \textbf{Inference Rule}& \inferenceRule \\
            \bottomrule
        \end{tabular*}
        \egroup
        \caption{A deductive parsing algorithm for \emph{Template}. Each item encodes how a computation (encoded by a tree of relations $t\in\texttt{Tree}_R$) corresponds to its computation type $\texttt{Type}_X$, and its interval type $\texttt{Type}_I$. The computation type defines the composibility of \emph{relations} whereas the interval type defines the composibility of \emph{surface segments}. The functions involved in the consequent of \textsc{complete-rep} are faillable (monadic computation) and the inference rule matches only when no function fails. } 
    \end{table*}
    
\subsection{Implementation}

    The weighted deductive system is carried out via a generic evaluation strategy called ``forward chaining,'' where the set of proven statements (chart) is expanded by repeatedly combining items using inference rules  until nothing can be deduced anymore. The evaluated deductive system produces a directed hyper-graph that compactly represents all the possible derivations of the sequence. A careful implementation can potentially make weighted deduction to be as efficient as the unweighted equivalent (e.g. recognition) \parencite{Jason2023time}. 
    To keep the algorithm tractable, we restrict the relations' arity 
    to no more than two as it is the minimal case where repeat can happen both among siblings and between parent and child (recursion) in a computational flow. Since we are looking for the minimal weighted item, Knuth's generalization of Dijkstra's algorithm can be used to prune the search space \parencite{knuth1977generalization}.

    For a given sequence, the set of minimal template programs that generates the sequence are obtained by traversing the produced hypergraph backward, each times selecting the hyperedges that result in the minimal weight. This process would produce a subgraph of the hypergraph compactly representing all the template programs that achieves the minimal size. 

        \newcommand{\coffeehisto}{
        \centering
        \fontsize{6}{10}
        \includegraphics[width=\textwidth]{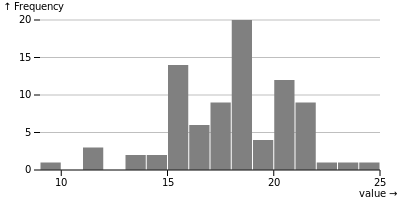}
        \caption{}
        \label{coffeehisto}
    }

    \newcommand{\chordhisto}{
        \centering
        \fontsize{5}{10}
        \includegraphics[width=\textwidth]{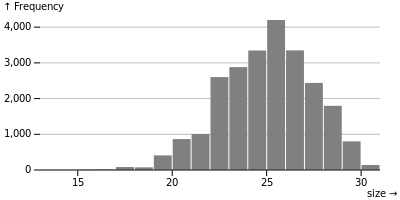}
        \caption{}
        \label{chordhisto}
    }

    \newcommand{\coffeeParse}{
        \centering
        \includegraphics[width=1\linewidth]{figures/coffeeParse.png}
        \caption{}
        \label{fig:coffeeParse}
    }
    \newcommand{\coffeeResult}{
        \centering
        \includegraphics[width=0.65\linewidth]{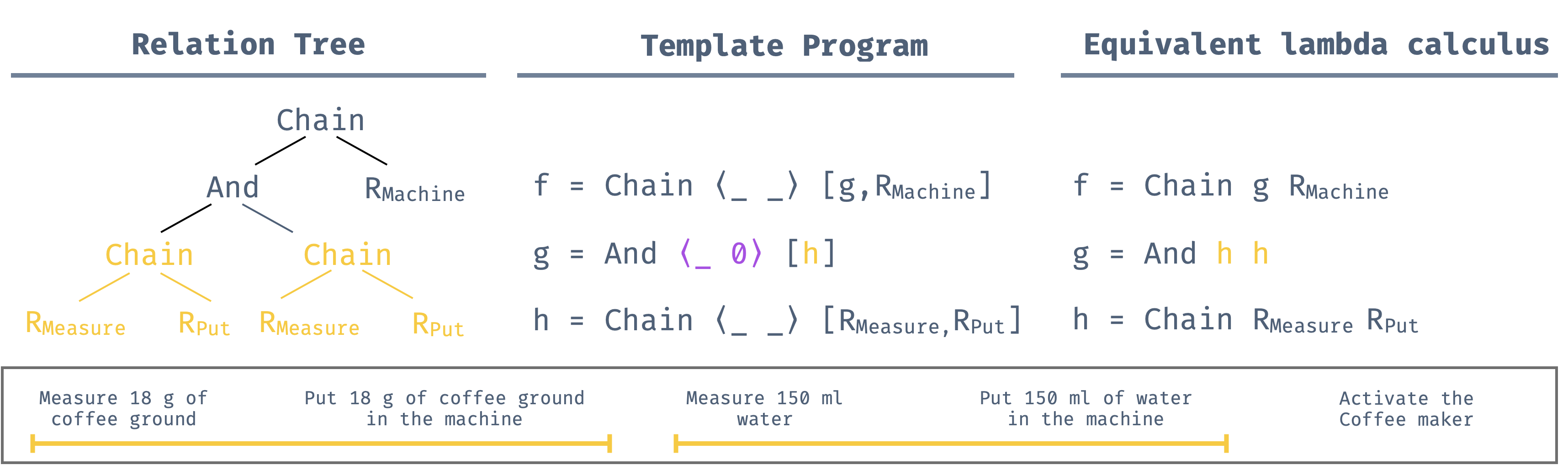}
        \caption{}
        \label{fig:coffeeResult}
    }
    \newcommand{\chordResult}{
        \centering
    \includegraphics[width=0.65\linewidth]{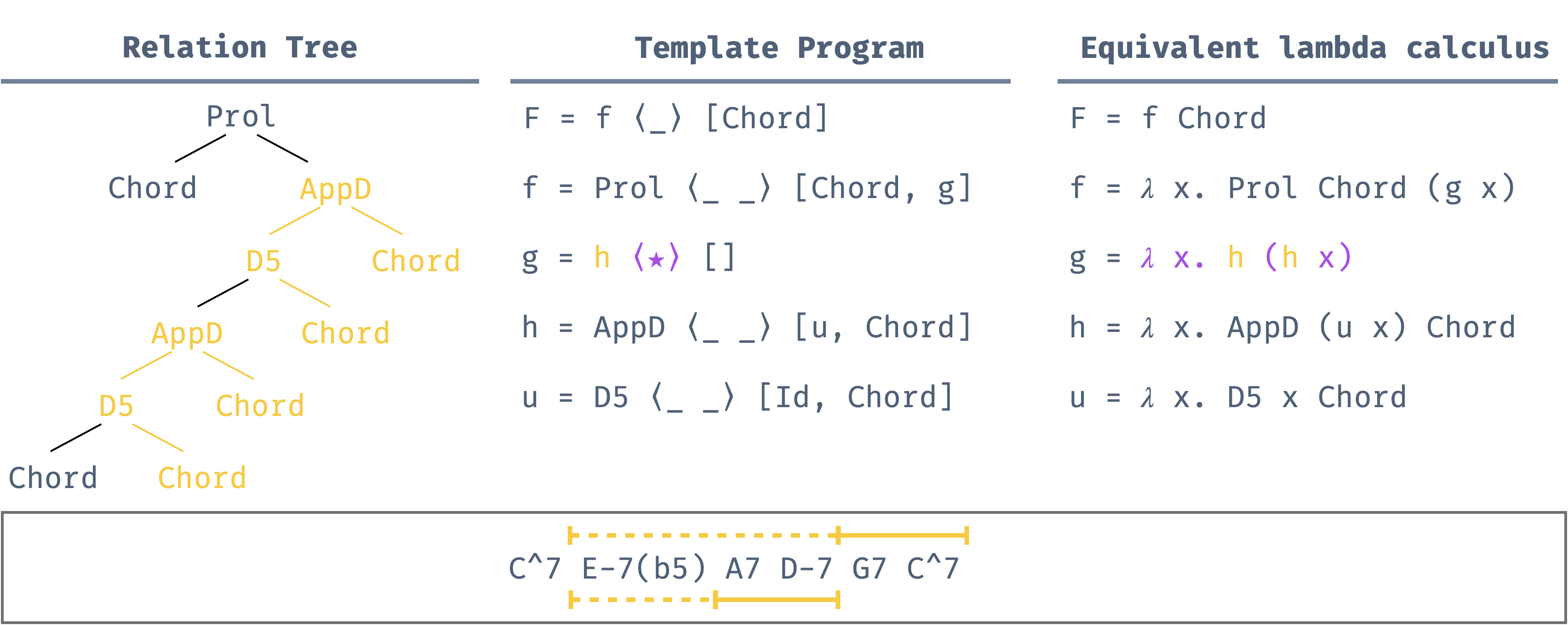}
        \caption{}
        \label{fig:chordResult}
    }
    
    \begin{figure*}[t]
        \begin{subfigure}[]{0.3\textwidth}
            \centering
            \begin{subfigure}[t]{\columnwidth}
                \coffeehisto
            \end{subfigure}
            \vfill
            \begin{subfigure}[b]{\columnwidth}
                \chordhisto
            \end{subfigure}
            \label{twohistograms}
        \end{subfigure}
        ~
        \begin{subfigure}[]{0.8\textwidth}
            \begin{subfigure}[t]{\linewidth}
                \coffeeResult
            \end{subfigure}
            \vfill
            \begin{subfigure}[b]{\linewidth}
                \chordResult
            \end{subfigure}
        \end{subfigure}
        
        \caption{Results for inferring the minimal template program in two contrasting domains: action planning and music. Fig. (\ref{coffeehisto} and \ref{chordhisto}) shows the template size distribution for the coffee action planning and jazz chord progression. Fig. (\ref{fig:coffeeResult} and  \ref{fig:chordResult})  The inferred minimal template program for action planning in making coffee reveals hidden yet cognitively plausible relational repeat within the sequential data.}
        \label{fig:results}
    \end{figure*}

\section{Proof-of-concept Demonstration} 

    We demonstrate our model on two minimal example sequences in the domain of action planning and music. Our objectives are twofold: first, to assess whether template programs can effectively capture the relational repeats in these sequences, and second, to evaluate the effectiveness of minimal description length as a heuristic for disambiguation in these specific examples. 
    For each of the two domains, we first encode domain-specific relations—such as those illustrated on the right side in Fig. \ref{fig:coffee} for the coffee making example—as primitive relations for the template program. For the chord sequence, our primitive relations\footnote{ ``AppD'' stands for applied dominant relation. ``D5'' stands for descending fifth relation. ``Prol'' stands for prolongation. ``Chord' is the termination rule.} are production rules from a simplified version of a jazz harmony grammar \parencite{rohrmeier2020syntax}.    This defines a constrained and interpretable hypothesis space specific to each domain. Subsequently, the model generates a minimal template program that explains the relational structure of each sequence.
    
    The inferred minimal programs and their encoding relations are shown in Fig. \ref{fig:results}. The minimal template program size is 9 for the coffee example and 13 for the chord progression example. To put these minimum size in context, we report the distribution of all possible template program sizes for the two sequences (see Fig. \ref{coffeehisto} \& \ref{chordhisto}). Even for a single parse tree, there exist many template programs that generate the same computation flow. This creates enormous ambiguous results in the high number of total template programs.
    
    In both examples, the minimal templates correspond to a meaningful and efficient description of the relations underlying the sequence. Unlike context-free grammar, whose generation process can be summarized as applications of independently sampled production rules, the template program can derive the same relation tree in a much shorter way. In the coffee example, the combinator $\langle \_ \ 0 \rangle$ corresponds to duplicating the template $h$, which translates to the two yellow portions of the relation tree and two surface segments (see Fig. \ref{fig:coffeeResult}). These surface segments are particularly interesting because the discovered repetition of relation aligns with humans intuitions. It is highly plausible that we abstract the combination of these two actions as preparing ingredients for the coffee recipe (``measure y amount of x and put that in the machine'').  
    The jazz chord progression provides an even more interesting result (see Fig. \ref{fig:chordResult}). The repeated relation is not a complete computation (i.e., tree containing holes). This means that it discovered a function-like structure that can take different inputs. 
    This minimal template captures recursion (repeated relation between parent and child computation) as shown by the usage of combinator $\langle \star \rangle$. This repeated structure is commonly known as ``ii-V-I'' and is widely used in jazz music. 

\section{Discussion}

    In this work, we developed a cognitively plausible computational model to explain how humans extract abstract repeated patterns from sequential data. We use Template programs as our hypothesis space. A core feature of the Template programs is the explicit duplication of dynamically built relations both among siblings and parent/child within a computation flow. 

    We presented an algorithm in the form of a weighted deduction system to infer the smallest template program that explains the relational structure of sequential data. 
    Finally, we demonstrated our model's capability in finding the efficient and cognitively plausible explanations for the observed sequences in music and action planning. These two case studies serve as a proof of concept of the broader potential for finding repetition structures in different domains.

    Despite our model's expressive power, finding the smallest template program is currently computationally impractical for longer sequences. In the future, we plan to further optimize our parsing algorithm and its implementation.
    One direction is to develop Earley style parser \parencite{earley1970efficient} which can binarize the inference rules and reduce unpromising items in the chart. 
    
    Our computational model can be applied to study human pattern recognition in different domains, offering flexibility and control over the space of relations. It has important implications for future cognitive psychological and neuroscientific studies of human pattern recognition. By focusing on the abstract relations governing sequential data—rather than surface-level features—it provides a framework for generating empirically testable predictions about how humans infer and represent underlying patterns.

\printbibliography

\end{document}